\documentclass[letterpaper, 10 pt, conference]{ieeeconf}
\IEEEoverridecommandlockouts
\usepackage{amsmath,amssymb,amsfonts}
\usepackage{bm}
\usepackage{graphicx}
\usepackage{cite}
\usepackage{tikz}
\usepackage{booktabs}
\usepackage{url}
\usepackage{wrapfig}
\usetikzlibrary{positioning,arrows.meta,fit}

\newcommand{\vect}[1]{\bm{#1}}
\newcommand{\mat}[1]{\bm{#1}}
\newcommand{\extendedWidth}{31.3}
\newcommand{\foldedWidth}{14.1}
\newcommand{\minimumGapWidth}{15}
\newcommand{\widthReductionPercentage}{55}

\usepackage{titlesec}
\usepackage[hidelinks]{hyperref}
\usepackage{xcolor}

\titlespacing*{\section}
{1pt}{1.2ex plus .5ex minus .2ex}{0.8ex plus .2ex}

\titlespacing*{\subsection}
{0pt}{0.85ex plus .3ex minus .2ex}{0.6ex plus .2ex}

\titlespacing*{\subsubsection}
{1pt}{0.8ex plus .2ex}{0.5ex}

\titlespacing*{\paragraph}
{1pt}{0.8ex}{0.5em}

\begin{document}
\bstctlcite{IEEEexample:BSTcontrol}

\title{\LARGE \bf Task-Driven Closed-Loop Design Optimization of Passively Morphing Quadrotors for Contact-Mediated Narrow Gap Traversal}

\author{Etor Arza$^{1}$, Luca Girardi$^{2}$, Gabriel Maquignaz$^{2}$, Javier Paez$^{1}$,\\ Stefan Leutenegger$^{2}$, Stefano Mintchev$^{3}$, and Kostas Alexis$^{1}$%
\thanks{$^{1}$Department of Engineering Cybernetics, Norwegian University of Science and Technology (NTNU), Trondheim, Norway. Corresponding author: Etor Arza, {\tt\small etora@ntnu.no}}%
\thanks{$^{2}$Mobile Robotics Lab, Department of Mechanical and Process Engineering, ETH Zurich, Zurich, Switzerland.}%
\thanks{$^{3}$Agricultural Robotics, University of Bonn, Bonn, Germany.}%
}

\maketitle
\thispagestyle{empty}
\pagestyle{empty}

\begin{abstract}  %
Aerial robots operating in cluttered environments may encounter openings narrower than their nominal width.
Existing passively morphing drones reduce their width upon traversal, but they typically involve human-made designs, which can struggle to find the right balance among folding ratio, flight efficiency, and control tracking performance.
We introduce a closed-loop design method that jointly optimizes quadrotor geometry and the torque-deflection response of two passively folding arms driven by the success rate in simulation for contact-mediated gap traversal.
An optimization algorithm samples candidate designs that are evaluated in a contact-based simulation of the traversal task, where the task success rate is used as a feedback signal to sample progressively better designs over time.
The resulting prototype passively reduces its width from $31.3$ to $14.1$\,cm and traverses a $15$\,cm wide opening at a $2$\,m/s reference speed.
It also tolerates gap approach-angle errors of up to $30^{\circ}$ and folds through a $40$\,cm long confined passage.
Compared to a rigid twin, the soft configuration adds only $2$\,mm of position RMSE and less than $1\%$ electrical power during contact-free flight.
These results show that task-driven closed-loop design optimization can produce passively morphing airframes that are compact under contact, tolerant to approach error, and effectively rigid during nominal flight.
\end{abstract}

\section{Introduction}

The width of an aerial robot can impose a hard limit on the environments it can access. 
In cluttered indoor, industrial, and natural settings, an otherwise traversable route may contain openings narrower than the vehicle's nominal width. 
Rigid quadrotors can reduce their projected cross-section through aggressive attitude maneuvers~\cite{falangagap,loianno}, but the opening must still exceed the vehicle's minimum projected dimension. 
Moreover, successful traversal requires accurate perception, state estimation, and timing, making this strategy difficult to deploy reliably in real-world settings.

Morphing aerial robots can overcome this geometric limitation by temporarily reducing their width.
For instance, actuated morphologies~\cite{quadmorphing,foldable,dragon} can decrease width on demand, but increase system complexity through additional servomotors and adaptive control strategies. 
Thrust-triggered quadrotor arm folding through passive joints~\cite{bucki2019,bucki2023} can achieve large and rapid shape changes, but requires invertible thrust and limiting maneuverability to avoid unintentionally triggering the passive joints. 
Contact-mediated passively reconfigurable designs are therefore particularly attractive as they leverage contact with the environment to initiate and maintain shape change~\cite{squeeze,morphy,girardi2025bioinspired,fabris2021soft,fabris2022crash} without requiring additional actuators or invertible thrust.
However, the geometry and compliance of these existing designs are not specifically optimized for the traversal task, where the trade-off between in-flight rigidity (and consequently tracking accuracy and power consumption) and the maximum achievable squeeze is not specifically addressed. 
Aerial robots that achieve contact-mediated in-flight narrow-gap traversal only pass through gaps up to 29\% narrower than the drone's span~\cite{squeeze,morphy,girardi2025bioinspired}.

A well-designed passively morphing quadrotor must remain effectively rigid under rotor loads during nominal flight and simultaneously easily reduce its width when the opening applies contact forces.
After traversal, it must restore the nominal flight configuration. 
A design that is too flexible can deform during normal flight, leading to worse control, while a stiffer structure can be difficult to compress through the gap.
When airframe geometry and compliance are selected by hand, it is particularly difficult to identify which combinations best enable both successful gap traversal and good nominal flight control performance.
Previous works show contact-mediated narrow-gap traversal is possible but generally propose a planar quadrotor architecture and compliance guided by engineering intuition.
To the best of our knowledge, no prior work has jointly optimized airframe geometry and compliance directly driven by closed-loop performance over the complete narrow-gap traversal and recovery task.

\begin{figure*}[t]
  \centering
  \includegraphics[width=\textwidth]{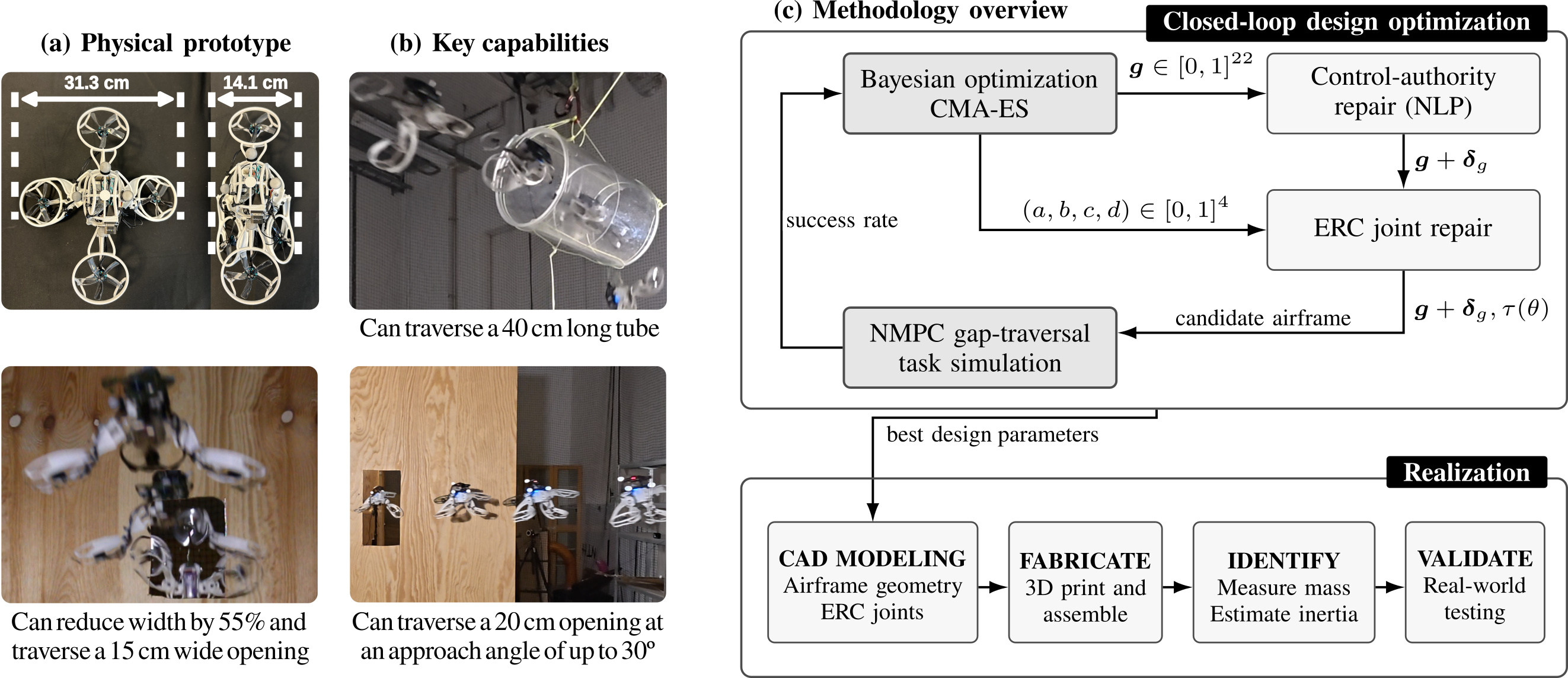}
  \caption{Task-driven closed-loop design optimization of passively morphing quadrotors for contact-mediated narrow-gap traversal.
  (a) Physical prototype in extended and folded configurations, with width reduced from $\extendedWidth$ to $\foldedWidth$\,cm.
  (b) Airframe capabilities: contact-induced arm folding enables traversal through a long tube and narrow openings, including angled approaches.
  (c) Airframe geometry and passive joint responses are jointly optimized using simulated gap-traversal success rate, with repairs for control authority and joint realizability, followed by fabrication and real-world validation.}
  \label{fig:overview}
\end{figure*}

In this work, we introduce a task-driven closed-loop design optimization approach for soft quadrotors, optimized for narrow-gap traversal.
Figure~\ref{fig:overview} summarizes the proposed closed-loop design optimization pipeline and the resulting vehicle's key capabilities.
We jointly optimize a 22-parameter quadrotor geometry and the programmable nonlinear torque deflection response of two passively folding side arms directly for traversal success.
The compliant response is achieved using elastic rolling cam (ERC) joints~\cite{erc}, whose cam profiles can be synthesized to produce nonlinear restoring torque as a function of the bending angle. 
This nonlinearity can be tuned to allow airframes to remain rigid near the nominal configuration, while bending with low effort after contact.
A black-box optimizer evaluates candidate designs using GPU-parallel contact simulation with a nonlinear model predictive controller (NMPC) whose model parameters are instantiated analytically from each candidate morphology.
Control-authority and joint-realizability repair procedures modify infeasible candidates before evaluation, ensuring that the search considers only manufacturable designs that retain at least $95\%$ of the acceleration capabilities (maximal angular acceleration in roll, pitch, and yaw, in addition to thrust to weight) of a planar reference design in + configuration that shares Morphy's arm length and propulsion system~\cite{morphy}.
The proposed approach allows the complete \emph{contact} $\rightarrow$ \emph{folding} $\rightarrow$ \emph{recovery} sequence to select a task-adapted design, without relying on geometric or mechanical proxy objectives.
An ablation study shows that this joint optimization is key, as optimizing geometry or joint compliance alone produces designs that fail at narrow gaps, whereas optimizing both together produces a design that traverses gaps $52\%$ narrower than its nominal width.
The complete pipeline produces a flight-ready design in about $18$\,h on a single laptop GPU.
The accompanying video is available at \textcolor{blue}{\url{https://youtu.be/xGkEtl6-vCA}}.
The code is available at \textcolor{blue}{\url{https://github.com/ntnu-arl/soft-quadrotor-optimization}}.

The main contributions of this work are:
\begin{itemize}
  \setlength{\itemsep}{1pt}
  \item A closed-loop design optimization method that jointly optimizes the geometry and passive joint compliance of a morphing quadrotor directly from simulated gap-traversal success rate.
  \item A real-world prototype that passively folds from a $31.3$ to a $14.1$\,cm width, and traverses a $15$\,cm-wide opening ($52\%$ narrower than its unfolded width) and a $40$\,cm-long, $22.5$\,cm-diameter tube ($28\%$ narrower).
  \item Successful traversal of a $20$\,cm gap ($36\%$ narrower) with approach angle errors of up to $30^{\circ}$, demonstrating tolerance to misaligned contact.
  \item Compliance-mediated passive morphing that adds only $2$\,mm of position RMSE and less than $1\%$ electrical power relative to a rigid twin in nominal flight.
\end{itemize}

\section{Methodology}
\label{sec:methodology}

Each of the considered designs is a quadrotor in + configuration with two rigid arms (front, back) and two soft side arms that fold passively under lateral contact (see Figure~\ref{fig:overview}a).
We optimize the geometry of a quadrotor, the orientation of the joint axes of its two soft foldable arms, and the elastic response of its arms such that the vehicle can fly through the narrowest gap possible.
We use the platform scale of Morphy~\cite{morphy} as a case study.
Our reference design is a planar quadrotor with Morphy's propulsion system (iFlight XING 1404 3000\,KV motors with HQProp Duct-T75 five-blade propellers) and arm length ($12.3$\,cm from the center to each motor), but in + configuration.

Figure~\ref{fig:overview}c summarizes the optimization pipeline.
The optimization algorithm samples a real valued vector of 22 geometry parameters and 4 joint response parameters that represent a candidate design to be evaluated.
The sampled design is not necessarily realizable, so in a first step, we repair the candidate design 1) to ensure good control authority (an angular acceleration capability and thrust-to-weight) close to those of a planar reference design and 2) to ensure physically realizable joint responses.
Repairing a design involves finding the closest possible design that meets the criteria above.
The repaired design is evaluated on the gap-traversal task in GPU-parallel simulation using a nonlinear model predictive controller (NMPC).
Its success rate over a fixed set of gap widths is returned to the optimization algorithm, which provides the next candidate design.
After optimization, the best design is modeled in CAD, 3D printed, assembled, and tested in the real world.
In the rest of this section, we explain this pipeline in more detail.

\subsection{Elastic Rolling Cam Joints}
\label{sec:erc_joints}

An elastic rolling cam (ERC) joint is a passive joint with a single degree of freedom and a programmable torque-deflection response~\cite{erc}.
It consists of two spring-loaded cams that roll against each other, with interlocking teeth that prevent slip.
As the joint bends, the cam profiles alter the spring extension and the joint's moment arm, producing a restoring torque $\tau(\theta)$ that depends on the joint deflection $\theta$.
The cam geometry thus defines the passive nonlinear elastic response without requiring a dedicated actuator.
Through inverse design, cam profiles can be generated from a desired torque-deflection curve, as long as the resulting geometry is convex and the spring stroke and energy limits are satisfied~\cite{erc}.

\begin{figure}[t]
  \centering
  \begin{minipage}[t]{0.49\columnwidth}
    \centering
    {\footnotesize (a) ERC joint response}\\[1mm]
    \includegraphics[width=\linewidth]{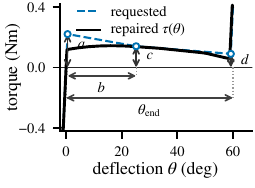}
  \end{minipage}\hfill
  \begin{minipage}[t]{0.49\columnwidth}
    \centering
    {\footnotesize (b) Geometry parameterization}\\[1mm]
    \includegraphics[width=\linewidth]{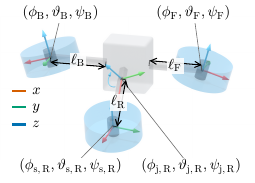}
  \end{minipage}
  \caption{(a) Requested (dashed) and repaired (solid) ERC restoring-torque response parameterized by $(a,b,c,d)$. (b) Geometry parameters on the optimized airframe.}
  \label{fig:erc_geometry}
\end{figure}

For each side arm, the self-collision angle $\theta_{\mathrm{end}}$ defines the maximum admissible folding angle before the arm collides with another part of the airframe.
We parameterize the target torque-deflection response of each ERC joint with four variables $(a,b,c,d)$ that define a piecewise-linear profile over the folding range $[0,\theta_{\mathrm{end}}]$ (Figure~\ref{fig:erc_geometry}a).
The parameter $a$ is the torque reached at the end of a short initial stiff segment, $c$ is the torque at an intermediate point placed at an angle proportional to $b$, and $d$ is the torque just before the maximum bending angle $\theta_{\mathrm{end}}$.
Deflections outside of this range are blocked with a stiff response.

Not every desired torque response can be obtained with an ERC joint (there are energy, cam convexity and spring stroke constraints)~\cite{erc}.
Therefore, a joint-response repair procedure maps a requested (desired) torque response to the closest physically realizable (repaired) response $\tau(\theta)$.
We additionally require the response torque to reach, within the first degree of deflection, at least $110\%$ of the maximum torque that the arm's rotor can exert about the joint axis at full thrust.
This 10\% margin keeps the arm rigid under rotor loads during nominal flight.
The repaired response is implemented as a restoring-torque lookup table in simulation.

\subsection{Soft Airframe with Programmable Joints}
The robot morphology is a quadrotor in + configuration with two rigid arms (front, back) and two soft side arms.
Each side arm consists of a short base segment and an outer segment that carries the motor and is connected through a passive ERC joint.
Each candidate design is encoded as a real valued vector $\vect{x} = (\vect{g},a,b,c,d) \in [0,1]^{26}$, where $\vect{g}$ contains the 22 airframe-geometry parameters and the four scalars $(a,b,c,d)$ that define the torque-deflection requested as explained before.

The geometry parameters $\vect{g}=(g_1,\ldots,g_{22})\in[0,1]^{22}$ are listed in Table~\ref{tab:geometry_parameters} with each component mapped affinely from $[0,1]$ to its listed physical range (see Figure~\ref{fig:erc_geometry}b for an illustration).
These geometry parameters also determine the position and orientation of the axis about which each side arm folds.
We limit the allowed joint axis orientations so that contact forces from the gap edges act primarily in the direction that bends each side arm about its joint axis, causing it to fold.
Large forces acting in other directions would bend or twist the joint in ways it is not designed to accommodate, making folding less reliable and increasing the risk of arm dislocation.
The optimized column reports the repaired geometry $\vect{g}+\vect{\delta}_g$ used in simulation and hardware.
Rotating a propeller about its own thrust axis does not change its thrust or drag torque.
We therefore map each propeller orientation to an equivalent canonical orientation with the same thrust direction and the yaw angle closest to zero, which is why all propeller yaw angles in Table~\ref{tab:geometry_parameters} are zero.

\begin{table}[t]
  \centering
  \caption{Geometry parameter ranges and optimized values.}
  \label{tab:geometry_parameters}
  \scriptsize
  \setlength{\tabcolsep}{1.5pt}
  \begin{tabular}{@{}c l c c c@{}}
    \toprule
    Variables & Parameter & Symbols & Range & Optimized \\
    \midrule
    $g_1,g_2$       & Front/Back arm length          & $\ell_{\mathrm F},\ell_{\mathrm B}$                         & $[5,15]$\,cm       & $\phantom{-}8.51\phantom{-}/\phantom{-}8.99$ \\
    $g_3,g_4$       & Left/Right arm length        & $\ell_{\mathrm L},\ell_{\mathrm R}$                         & $[5,15]$\,cm       & $\phantom{-}8.89\phantom{-}/\phantom{-}8.98$ \\
    $g_5,g_6$       & Front/Back propeller roll      & $\phi_{\mathrm F},\phi_{\mathrm B}$                         & $[-45,45]^\circ$   & $\phantom{-}0.42\phantom{-}/\phantom{-}0.44$ \\
    $g_7,g_8$       & Front/Back propeller pitch     & $\vartheta_{\mathrm F},\vartheta_{\mathrm B}$               & $[-45,45]^\circ$   & $\phantom{-}16.57\phantom{-}/\phantom{-}17.21$ \\
    $g_9,g_{10}$    & Front/Back propeller yaw       & $\psi_{\mathrm F},\psi_{\mathrm B}$                         & $[-45,45]^\circ$   & $\phantom{-}0.00\phantom{-}/\phantom{-}0.00$ \\
    $g_{11},g_{12}$ & Joint axis roll     & $\phi_{\mathrm{j,L}},\phi_{\mathrm{j,R}}$                   & $[-30,30]^\circ$   & $-24.15\phantom{-}/\phantom{-}30.00$ \\
    $g_{13},g_{14}$ & Joint axis pitch    & $\vartheta_{\mathrm{j,L}},\vartheta_{\mathrm{j,R}}$         & $[-45,45]^\circ$   & $\phantom{-}37.02\phantom{-}/\phantom{-}37.73$ \\
    $g_{15},g_{16}$ & Joint axis yaw      & $\psi_{\mathrm{j,L}},\psi_{\mathrm{j,R}}$                   & $[-5,5]^\circ$     & $\phantom{-}1.39\phantom{-}/\phantom{-}3.85$ \\
    $g_{17},g_{18}$ & Left/Right propeller roll     & $\phi_{\mathrm{s,L}},\phi_{\mathrm{s,R}}$                   & $[-45,45]^\circ$   & $\phantom{-}22.47\phantom{-}/-28.51$ \\
    $g_{19},g_{20}$ & Left/Right propeller pitch    & $\vartheta_{\mathrm{s,L}},\vartheta_{\mathrm{s,R}}$         & $[-45,45]^\circ$   & $-19.55\phantom{-}/-21.36$ \\
    $g_{21},g_{22}$ & Left/Right propeller yaw      & $\psi_{\mathrm{s,L}},\psi_{\mathrm{s,R}}$                   & $[-45,45]^\circ$   & $\phantom{-}0.00\phantom{-}/\phantom{-}0.00$ \\
    \bottomrule
  \end{tabular}
\end{table}

\subsection{Control Authority Repair}
\label{sec:repair}
In this work, we define \textit{control authority} as the maximal angular acceleration and vertical thrust acceleration that the vehicle can achieve while compensating for gravity.
We repair candidates with insufficient control authority by solving for the smallest parameter change $\vect{\delta}_g$ that restores the required acceleration capabilities relative to a reference planar design.
Specifically, we require the repaired design $\vect{g}{+}\vect{\delta}_g$ to retain at least $95\%$ of the reference design's maximum vertical thrust acceleration (with zero lateral force and torque), denoted as $a_1$, positive and negative roll ($a_2,a_3$) and pitch ($a_4,a_5$) angular accelerations while compensating gravity, and positive and negative yaw angular accelerations ($a_6,a_7$).
Each capability $a_j$ is satisfied by motor forces $\vect{f}_j$ acting through the allocation matrix $\mat{B}(\vect{g}{+}\vect{\delta}_g)$.
Denoting the reference capability values by $\bar{a}_j$, we formulate the repair as
\begin{equation}
\label{eq:repair}
\begin{aligned}
\min_{\vect{\delta}_g,\, \vect{f}_1,\dots,\vect{f}_7} \quad & \lVert \vect{\delta}_g \rVert_2 \\
\text{s.t.} \quad & a_j(\vect{g}{+}\vect{\delta}_g, \vect{f}_j) \ge (0.95)\, \bar{a}_j,
\quad j = 1,\dots,7, \\
& \vect{f}_j \in [f_{\min}, f_{\max}]^4, \qquad \vect{g}+\vect{\delta}_g \in [0,1]^{22},
\end{aligned}
\end{equation}
where $f_\text{min},f_\text{max}$ are the minimum and maximum motor forces.

\subsection{Gap-Traversal Task}
We evaluate the candidate designs in Isaac Lab~\cite{isaaclab} with GPU-parallel rigid-body physics at $500$\,Hz and control at $100$\,Hz.
Each of the $100$ parallel environments contains a wall with a rectangular gap, with gap widths evenly spaced over $[0.15, 0.25]$\,m.
The $25$\,cm upper bound is set just below Morphy's $25.2$\,cm nominal width~\cite{morphy}.
The lower end of the gap size is above the $12$\,cm minimum folded width (set by the $5$\,cm-wide electronics cage and each side arm's $2$\,cm fixed base segment and $1$\,cm-thick folding segment per side).

In each episode, the vehicle starts in hover $2$\,m in front of the wall and tracks a reference that crosses the gap center at a forward speed of $2.0$\,m/s and ends in hover $1$\,m behind the wall.
The side arms contact the gap edges and fold, the vehicle squeezes through, and the arms re-extend under their elastic response.
An episode is a \emph{success} if the vehicle reaches $0.6$\,m beyond the wall with a speed below $0.25$\,m/s and with both arms re-extended to within $3^{\circ}$ from the nominal configuration.
It is a \emph{crash} if the vehicle is blocked by the wall, exceeds the attitude-error limits, leaves the flight volume, or stalls inside the gap, and a \emph{timeout} if it fails to succeed in 2.5\,s.
To encourage safe traversal, attitude errors above $30^{\circ}$ near the gap count as a crash during optimization even when the vehicle could recover.

The goal of the optimization is to find a design that has a high traversal success rate at every gap width.
Estimating success rate accurately would be computationally costly, with $400$ rollouts per gap width for example requiring approximately $9$ hours per design.
We therefore maximize a cheaper proxy, calculated as the percentage of successful episodes over the $100$ environments with one rollout per width, which takes about $80$\,s per design.

\begin{figure}[t]
  \centering
  \includegraphics[width=\columnwidth]{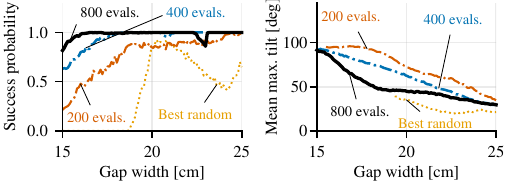}
  \caption{Repeated evaluation of four designs with $400$ trials per gap width.
  Left: success probability.
  Right: mean maximum tilt during the first 0.2 seconds after traversal among successful trials where success exceeds $30\%$.}
  \label{fig:objective_comparison}
\end{figure}

To check whether optimization using this proxy selects designs with high traversal success rates, we take the designs selected after $200$, $400$, and $800$ optimization evaluations and then reevaluate them alongside the best of $50$ random designs using $400$ rollouts per width and no attitude termination, as shown in Figure~\ref{fig:objective_comparison}.
As optimization progresses, the success rate across gap widths increases and mean maximum tilt decreases (we use the word \textit{tilt} to refer to the angle between the body vertical axis and the world vertical axis).

\subsection{NMPC controller}
Every design is flown by the same NMPC controller, with the model analytically defined from the candidate geometry.
The state $\vect{\xi} = (\vect{p}, \vect{v}, \vect{q}, \vect{\omega}, \vect{\Omega})
\in \mathbb{R}^{17}$ collects position, velocity, attitude quaternion, body rates, and the
four rotor speeds; the input $\vect{u}=(u_1,...,u_4) \in \mathbb{R}^{4}$ are commanded motor forces.
The dynamics are
\begin{equation}
\label{eq:model}
\begin{aligned}
\dot{\vect{p}} &= \vect{v}, \qquad
\dot{\vect{q}} = \tfrac{1}{2}\, \vect{q} \otimes (0, \vect{\omega}), \\
m \dot{\vect{v}} &= m \vect{g}_w + \mat{R}(\vect{q})\, \mat{B}_f \vect{f}, \\
\mat{J} \dot{\vect{\omega}} &= \mat{B}_\tau \vect{f} - \vect{\omega} \times \mat{J} \vect{\omega}, \\
\dot{\Omega}_i &= \big(\sqrt{u_i / k_{\mathrm{f}}} - \Omega_i\big)/t_{\mathrm{m}}, \qquad
f_i = k_{\mathrm{f}} \Omega_i^{2},
\end{aligned}
\end{equation}
where $\otimes$ denotes quaternion multiplication, $m$ is the total mass, $\mat{R}(\vect{q})$ is the body-to-world rotation matrix, $\vect{g}_w$ is the world-frame gravity vector, $\vect{f}{=}(f_1,\dots,f_4)$ is the vector of actual motor forces, $k_{\mathrm{f}}$ is the rotor thrust coefficient relating rotor speed to thrust ($f_i = k_{\mathrm{f}}\Omega_i^2$), $t_{\mathrm{m}}$ is the first-order motor time constant, and the allocation matrix $[\mat{B}_f\, \mat{B}_\tau]$, mass $m$, and inertia $\mat{J}$ are computed from the geometry parameters $\vect{g}+\vect{\delta}_g$ with the arms in their extended configuration.
Note that the NMPC model neglects the short-lived change in allocation matrix and inertia caused by ERC arm deflection during gap traversal, since this deflection is not measured and the arms quickly extend back after traversal.
The NMPC solves
\begin{equation}
\label{eq:ocp}
\begin{aligned}
\min_{\vect{u}_{0:N-1}} \;& \sum_{k=0}^{N-1} \vect{r}_k^\top \mat{W}\vect{r}_k
+ \vect{r}_N^\top \mat{W}_N\vect{r}_N \\
\text{s.t.}\;& \vect{u}_k \in [f_{\min}, f_{\max}]^4,
\end{aligned}
\end{equation}
subject to the dynamics in Equation~\eqref{eq:model}, with residual $\vect{r}_k = (\vect{e}_{p}, \vect{e}_{v}, \vect{e}_{R}, \vect{e}_{\omega}, \vect{u}_k, \Delta\vect{u}_k)$ containing errors in position, velocity, attitude rotation-vector, and body rates, plus input magnitude and input-change penalties, where the body-rate error $\vect{e}_{\omega} = \vect{\omega}$ uses a zero reference.
We use a prediction horizon of $0.5$\,s and solve it with the real-time iteration scheme of sequential quadratic programming in acados~\cite{acados}.

\begin{wraptable}{r}{0.43\columnwidth}
\centering
\vspace{2pt}
\refstepcounter{table}\label{tab:nmpc_weights}
{\footnotesize TABLE~\thetable: \textsc{NMPC weights.}}\par\vspace{4pt}
\footnotesize
\setlength{\tabcolsep}{1.5pt}
\begin{tabular}{@{}lcc@{}}
\toprule
& Flight & Recovery \\
\midrule
$\vect{e}_p$ & $(0.05{,}30{,}10)$ & $(2{,}2{,}10)$ \\
$\vect{e}_v$ & $(0.30{,}4{,}15)$ & $(3{,}3{,}5)$ \\
$\vect{e}_R$ & $(4{,}2{,}2)$ & $(4{,}2{,}2)$ \\
$\vect{e}_\omega$ & $(1{,}1{,}0.5)$ & $(1{,}0.5{,}0.5)$ \\
$\vect{u}$ & $0.12$ & $0.12$ \\
$\Delta\vect{u}$ & $35$ & $80$ \\
\bottomrule
\end{tabular}
\vspace{-2pt}
\end{wraptable}
The diagonal entries of $\mat{W}$ are the weights in Table~\ref{tab:nmpc_weights}, with triples ordered as $x,y,z$ and scalar input weights repeated for each motor.
We use two weight sets: a \emph{flight} set for gap approach and a \emph{recovery} set for stabilization.
During flight, the position weight along $y$ is $30$, compared with $0.05$ along $x$, to prioritize lateral alignment with the gap center.
Recovery reduces the lateral position weight to $2$, and increases the input-change penalty from $35$ to $80$ to avoid extreme motor commands that destabilize the drone.
The terminal residual $\vect{r}_N$ contains only the first four residual groups, and $\mat{W}_N$ is twice the corresponding diagonal block of $\mat{W}$.

For the real-world experiments, we further tuned some of the parameters.
In simulation, motor revolutions per second (rps) were limited between $83$ and $400$\,rps, as we experimentally verified that a first-order motor model fit the real motors well in this range.
For real-world experiments, we increased the lower bound to $200$\,rps and used $W_{\Delta u}=35$ during flight and $80$ during recovery (up from $W_{\Delta u}=25$ in simulation).
These parameters showed smoother motor actions and a better recovery after contact in the real drone.

\begin{figure*}[t]
  \centering
  \begin{minipage}[t]{0.325\textwidth}
    \centering
    \vspace{0pt}
    \includegraphics[width=\linewidth]{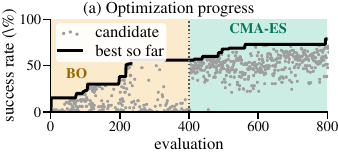}
  \end{minipage}\hfill
  \begin{minipage}[t]{0.325\textwidth}
    \centering
    \vspace{0pt}
    \includegraphics[width=\linewidth]{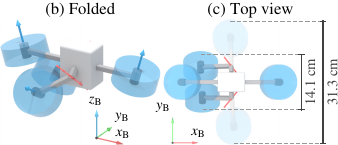}
  \end{minipage}\hfill
  \begin{minipage}[t]{0.325\textwidth}
    \centering
    \vspace{0pt}
    \includegraphics[width=\linewidth]{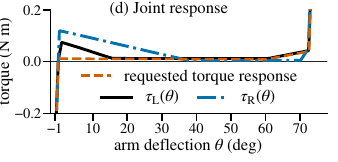}
  \end{minipage}
  \caption{Simulation optimization results. (a) Candidate success rates (gray), best-so-far curve (black), and BO--CMA-ES transition (dotted) over the $800$ evaluations. Optimized airframe (b) maximally folded and (c) overlaid with its extended configuration in top view; folding reduces its width from $\extendedWidth$ to $\foldedWidth$\,cm ($\widthReductionPercentage\%$). (d) Requested and repaired left/right ERC restoring-torque responses; positive $\theta$ is the folding direction and negative $\theta$ is blocked.}
  \label{fig:progress}
  \label{fig:design}
  \label{fig:optimized_torque_response}
\end{figure*}

\subsection{Optimization Strategy}
\label{subsec:optimization_strategy}
The optimization algorithm maximizes the success rate of the candidate airframes with a fixed budget of $800$ evaluations in two steps.
Each evaluation involves running the full repair--synthesize--simulate pipeline of Figure~\ref{fig:overview}c.
As in previous work on performance based multirotor optimization~\cite{taskspecific}, we first use Bayesian optimization (BO)~\cite{botorch} for $400$ evaluations, where the first $50$ are space filling initialization samples.
BO serves as a global exploration algorithm and is suitable for low-budget noisy optimization.
Then, covariance matrix adaptation evolution strategy (CMA-ES)~\cite{cmaes} performs local refinement around the best solution found by BO for another $400$ evaluations. %

\section{Evaluation Studies}
\label{sec:experiments}

\subsection{Simulation}
\label{sec:exp_optimization}

We ran the optimization pipeline described in Section~\ref{subsec:optimization_strategy}.
Each candidate is evaluated on the success rate over 100 parallel rollouts with gap widths evenly spaced in $[0.15, 0.25]$\,m.
Evaluating a design takes $80.8 \pm 6.7$\,s (mean $\pm$ standard deviation over all 800 designs), and the complete optimization finished in about 18\,h on a single laptop GPU (NVIDIA RTX 4080).

Figure~\ref{fig:progress}a shows the success rate of each candidate and the best success rate found so far.
The task is hard for random designs: the best of the $50$ space-filling samples that initialize BO succeeds in only $15\%$ of its rollouts.
BO identifies a promising region and raises the best success rate to $56\%$.
The subsequent local refinement with CMA-ES nearly halves the failure rate, and the final design reaches a success rate of $79\%$ ($21\%$ classified crashes, no timeouts)\footnote{The $21\%$ of crashes result solely from exceeding the strict $30^{\circ}$ attitude-error limit near the gap: if allowed to continue, every affected rollout recovers and meets the success conditions.}.
Many candidates in the CMA-ES phase obtain a high success rate, which indicates that the best design lies in a consistent well-performing region of the design space.

Figure~\ref{fig:erc_geometry}b shows the optimized airframe in its extended configuration, and Table~\ref{tab:geometry_parameters} reports its complete repaired geometry.
The airframe has approximate bilateral symmetry in its overall shape, where angular-acceleration capabilities differ by at most $0.24\%$ across roll, pitch, and yaw. 
We did not enforce bilateral symmetry in the parameterization, but the optimization still converged to an approximately symmetric design, likely due to symmetric control-authority constraints and because symmetric arm folding may facilitate gap traversal by minimizing yaw disturbances with a symmetric arm contact on the gap.

The obtained joint response has low requested restoring torques through most of the folding range (yield torque $a \approx 0.011$\,Nm and intermediate torque $c \approx 0.009$\,Nm at $83\%$ of the travel), followed by a rise to $d \approx 0.040$\,Nm near the end of travel (Figure~\ref{fig:optimized_torque_response}d).
The left and right repaired responses are intentionally different despite sharing the same requested profile because the rigidity constraint is evaluated independently for each arm.
The two side propellers rotate in the same direction, so the propeller drag torque opposes the thrust-induced joint moment on the left arm but reinforces it on the right.
For the optimized design, the resulting right-arm rigidity bound is about twice the left-arm bound ($0.116$ versus $0.057$\,Nm), producing the distinct repaired curves in Figure~\ref{fig:optimized_torque_response}d.
In successful simulated crossings, the arms remain extended before contact, fold while constrained by the gap, and re-extend after clearing it.
We build this design (Figure~\ref{fig:overview}a) and use it in the real-world experiments in the remainder of this section.

\subsubsection{Ablation Study}
\label{sec:ablation}

\begin{wrapfigure}{r}{0.48\columnwidth}
  \centering
  \vspace{-10pt}
  \includegraphics{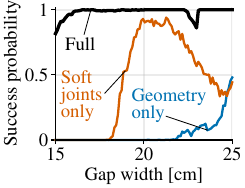}\par
  \vspace{-6pt}
  \caption{Success probability for the best designs optimized over the full and reduced design spaces.}
  \label{fig:design_space_ablation}
\end{wrapfigure}
We repeated the optimization of Section~\ref{subsec:optimization_strategy} with $800$ evaluations on two reduced design spaces.
The geometry-only variant optimizes $\vect{g}$ with rigid side arms.
The soft-joint-only variant fixes the airframe to a planar quadrotor and optimizes $(a,b,c,d)$ and the bending axes of the soft joints.
Figure~\ref{fig:design_space_ablation} shows the success probability per gap width.
Optimizing the geometry or the soft joints alone yields designs that fail at the narrower gaps.
Only when we optimize both together do we obtain a design that succeeds even at gaps $52\%$ narrower than its nominal width, whereas prior contact-mediated designs pass gaps at most $29\%$ narrower than their span~\cite{squeeze,morphy,girardi2025bioinspired}.

\subsubsection{Effect of the Optimization Seed}
\label{sec:seed_consistency}
\begin{figure}[t]
  \centering
  \includegraphics[width=\columnwidth]{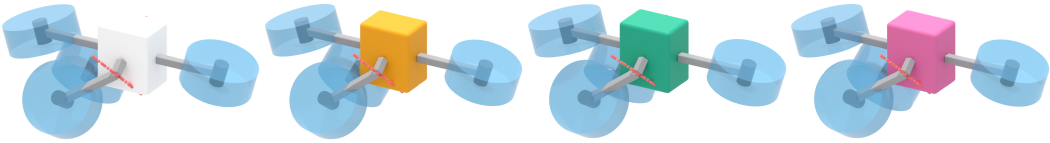}
  \includegraphics[width=0.85\columnwidth]{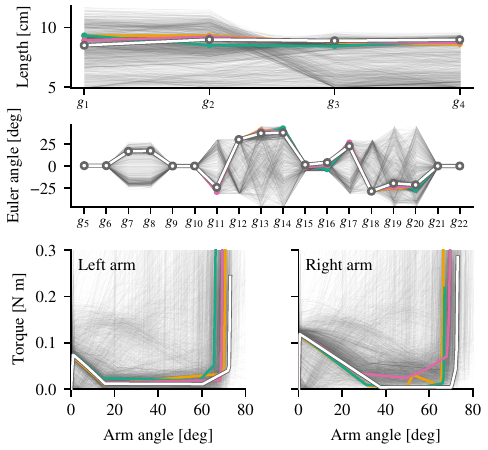}
  \caption{Best designs from four optimization runs with different seeds.
  Top: render of folded airframes.
  Below: geometry parameters and restoring torque.
  Colored lines show the parameters of the best found airframe for each seed (color matches render above), while gray lines show all evaluated airframes.}
  \label{fig:seed_consistency}
\end{figure}

We repeated the optimization with three additional seeds, for a total of four runs with $800$ candidate evaluations each.
The best designs from the four runs have similar arm lengths and orientations (see top of Figure~\ref{fig:seed_consistency}).
The repaired joint responses are similar across seeds for each arm, with a consistent difference between the left and right arms due to the rigidity constraints discussed above for the optimized design in Figure~\ref{fig:optimized_torque_response}d.
The top left design (white) is used in the physical experiments.

\subsection{Real-World Experiments}
\label{sec:exp_real}

\subsubsection{Gap Traversal}

We flew the vehicle in a motion-capture arena, where the motion-capture pose estimates are streamed to the onboard flight controller, and the NMPC described in Section~\ref{sec:methodology} runs onboard.
The experiment involves a wall with a $\minimumGapWidth \times 20$\,cm rectangular opening and a reference that crosses the gap center at $2$\,m/s and ends in hover behind the wall.
We report traversal positions relative to the gap center, where $x=y=z=0$.
The opening is less than half of the $\extendedWidth$\,cm rest (nominal) width of the vehicle and only slightly wider than its $\foldedWidth$\,cm fully folded width.
Therefore, the vehicle must fold almost completely to pass the gap in level flight.
For comparison, Morphy~\cite{morphy}, which uses the same propulsion system and a similar platform size, squeezes only from $25.2$ to $22$\,cm.

\begin{figure}[t]
  \centering
  \includegraphics[width=\columnwidth]{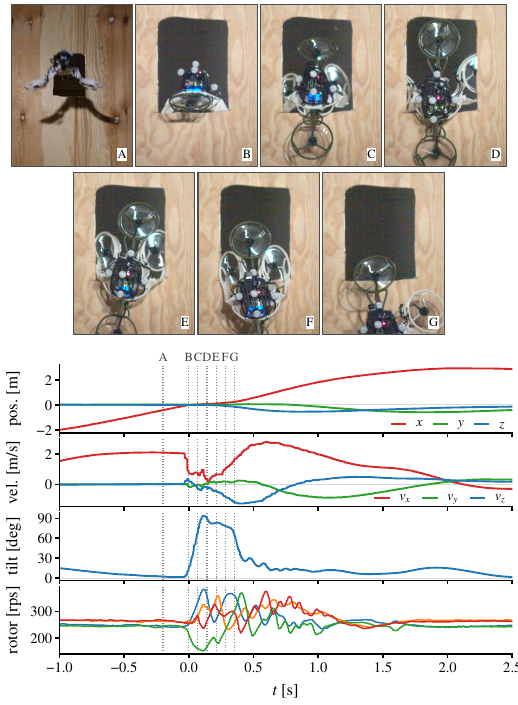}
  \caption{Traversal of a $15 \times 20$\,cm opening.
  Top: video frames of the crossing.
  Bottom: logged gap-relative position, velocity, tilt, and rotor speeds.
  Dotted vertical lines mark the frame instants.}
  \label{fig:real}
\end{figure}

Figure~\ref{fig:real} shows a traversal through this opening (video at \href{https://youtu.be/xGkEtl6-vCA?t=108}{\textcolor{blue}{1:48}}).
The vehicle approaches in level flight (frame~A) and reaches the gap well-centered: over the final half meter before contact, its maximum lateral deviation from the commanded crossing line is $5$\,mm and its mean forward speed is $2.07$\,m/s against the $2$\,m/s reference.
On contact, the gap edges fold both side arms, and the constrained squeeze rotates the vehicle to a peak tilt of $94^{\circ}$ within $0.11$\,s of the crossing (frames~B--D).
The vehicle slides down and bends the arms further (frames~D--F).
The arms re-extend under their elastic response after the wall traversal (frames~G and beyond).
The drone then recovers, and the vertical position reaches $z=-0.56$\,m while the controller prioritizes attitude recovery, and the tilt returns below $10^{\circ}$ within $2.2$\,s of the crossing.

\subsubsection{Traversal Through a Confined Tube}
\label{sec:exp_tube}

The wall experiment applies contact at a single plane, after which the arms are free to re-extend.
We further test the mechanical capability of the drone by replacing the wall with a transparent cylindrical tube aligned with the flight direction.
The tube is $40$\,cm long and $22.5$\,cm in diameter: its diameter is $72\%$ of the vehicle's $\extendedWidth$\,cm extended width.
For this experiment, the origin $x=y=z=0$ is at the center of the tube entrance.
The vehicle tracks the same $2$\,m/s crossing-speed reference towards the entrance of the tube, after which we add a fixed $0.25$\,s coast interval in which all four motors are commanded at $200$\,rps during the tube traversal.
The recovery NMPC then resumes.
This prevents the controller from fighting the tube constraint while the drone travels through the tube with the arms folded.

\begin{figure}[t]
  \centering
  \includegraphics[width=\columnwidth]{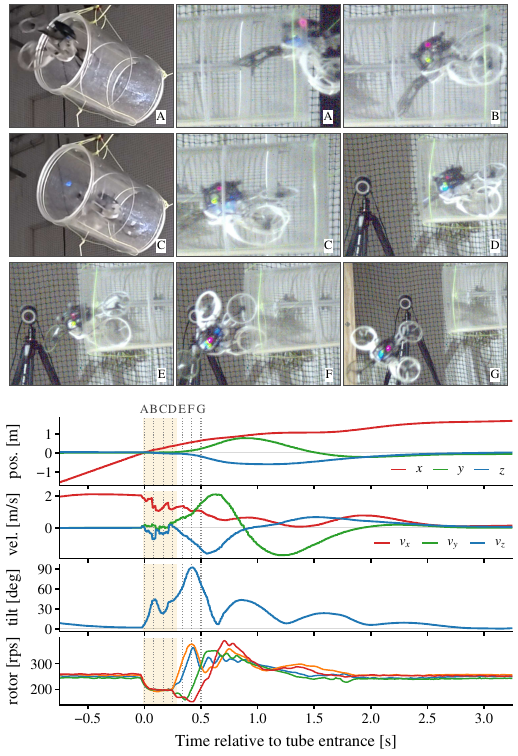}
  \caption{Traversal of a $40$\,cm-long, $22.5$\,cm-diameter tube.
  Top: video frames of the traversal.
  Bottom: entrance-relative position, velocity, tilt, and rotor speeds.
  The origin is at the tube entrance center where $x=y=z=0$.
  Dotted vertical lines mark the frame instants, and shading marks the interval in which the vehicle is inside the tube.}
  \label{fig:tube}
\end{figure}

Figure~\ref{fig:tube} shows the traversal (video at \href{https://youtu.be/xGkEtl6-vCA?t=134}{\textcolor{blue}{2:14}}).
Over the final half meter before the tube entrance, the maximum lateral deviation of the vehicle from the tube axis is $4$\,mm and its mean forward speed is $2.03$\,m/s.
Contact with the entrance folds both side arms (frames~A--B), and the tube holds them folded as the vehicle passes through (frames~C--E).
The vehicle center travels from the entrance to the exit in $0.30$\,s (area marked in yellow in the figure).
Once clear of the tube, the arms re-extend under their elastic response (frames~F--G) and the recovery NMPC recovers from the resulting motion.
During recovery, the vehicle reaches a peak tilt of $92^{\circ}$ and reaches $z=-0.60$\,m relative to the entrance center before stabilizing.

\subsubsection{Robustness and Approach Angle}
\label{sec:exp_angle}

\begin{figure}[t]
  \centering
  \includegraphics[width=\columnwidth]{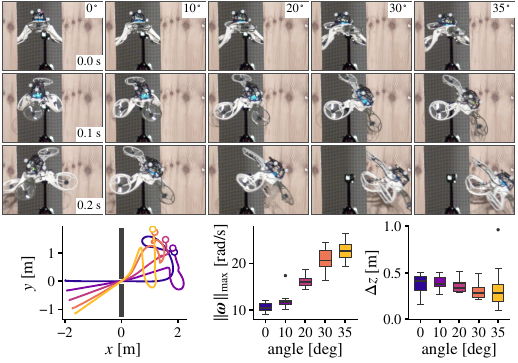}
  \par\vspace{-4pt}
  {\footnotesize Real-time video of all traversals: \textcolor{blue}{\url{https://youtu.be/TLzvmtZQMrI}}}\par\vspace{4pt}
  \caption{Traversal of a $20$\,cm opening at approach angles of $0$--$35^{\circ}$.
  Top: high-speed camera frames.
  Bottom: top view of the trajectory and boxplots over $20$ traversals per angle of the peak angular velocity and of the altitude loss $\Delta z$ on successful traversals.
  }
  \label{fig:angles}
\end{figure}

In a real deployment, the drone would rarely approach the gap perfectly aligned, yet prior morphing-airframe traversal experiments report only normally aligned approaches and do not test tolerance to approach-angle error~\cite{quadmorphing,foldable,dragon,bucki2019,bucki2023,squeeze,morphy,erc}.
In this experiment we further test the robot-and-controller solution which at design time only encountered aligned gap traversal, and test its robustness to gap orientation misalignment by repeating the traversal of a $20$\,cm-wide opening with the reference path rotated about the gap center by $0^{\circ}$, $10^{\circ}$, $20^{\circ}$, $30^{\circ}$, and $35^{\circ}$, keeping the vehicle, the controller, and the $2$\,m/s crossing reference speed unchanged (Figure~\ref{fig:angles}, video at \href{https://youtu.be/xGkEtl6-vCA?t=164}{\textcolor{blue}{2:44}}).
The drone flies $20$ consecutive traversals at each angle.

The drone geometry provides an ideal upper bound on the admissible approach angle.
Consider the fully folded drone as a strip of width $w_{\mathrm{fold}}$ aligned with its flight path, and approximate the opening as a plane, so that the two sides of the strip cross the wall plane $w_{\mathrm{fold}}/\cos\alpha$ apart at an approach angle $\alpha$.
Geometric passage through an opening of width $w_{\mathrm{gap}}$ requires $w_{\mathrm{fold}}/\cos\alpha\leq w_{\mathrm{gap}}$, giving $\alpha_{\max}=\cos^{-1}(w_{\mathrm{fold}}/w_{\mathrm{gap}})=45.2^{\circ}$ for $w_{\mathrm{fold}}=\foldedWidth$\,cm and $w_{\mathrm{gap}}=20$\,cm.

An angled approach breaks the symmetry of the contact: the leading arm touches its gap edge first, and the edge forces act along the flight direction rather than along the folding direction.
As a consequence, the contact impulse produces an increasingly large rotational disturbance.
The median peak angular velocity within $0.2$\,s of the crossing grows with the approach angle, from $10.5$\,rad/s at normal incidence, to $16.0$\,rad/s at $20^{\circ}$, and to $22.7$\,rad/s at $35^{\circ}$ (Figure~\ref{fig:angles}).

The leading side arm dislocated on contact with the wall in $1$ of $20$ traversals at $30^{\circ}$ and $3$ of $20$ at $35^{\circ}$ (Table~\ref{tab:failures_by_approach_angle}).
In all other traversals, the vehicle recovered to hover without loss of control, with a median altitude loss between $0.28$ and $0.41$\,m at each angle.
We count a drop of more than half of the $1.58$\,m gap-center height as an altitude-loss failure, which occurred once, with $0.96$\,m at $35^{\circ}$.
At $40^{\circ}$, the vehicle did not complete a successful traversal in multiple attempts.
Gap traversal is therefore robust on traversals up to $30^{\circ}$, which nears the ideal limit of $45.2^{\circ}$. 
Essentially, up to $30^{\circ}$, the system does not fail due to body design or control issues and only experiences mechatronic problems of arm dislocation owing to its prototype realization through 3D printing and other manufacturing limitations. At an approach angle of $35^{\circ}$, mechanical issues are more frequent, while one altitude loss-based failure is encountered.

\begin{table}[t]
  \centering
  \caption{Failures over $20$ consecutive traversals per approach angle. }
  \label{tab:failures_by_approach_angle}
  \footnotesize
  \begin{tabular*}{\columnwidth}{@{\extracolsep{\fill}}lccccc@{}}
    \toprule
    Approach angle & $0^{\circ}$ & $10^{\circ}$ & $20^{\circ}$ & $30^{\circ}$ & $35^{\circ}$ \\
    \midrule
    Arm dislocation & 0 & 0 & 0 & 1 & 3 \\
    Loss of control & 0 & 0 & 0 & 0 & 0 \\
    Altitude loss & 0 & 0 & 0 & 0 & 1 \\
    \midrule
    Max.\ altitude loss [m] & $0.50$ & $0.50$ & $0.51$ & $0.49$ & $0.96$ \\
    \bottomrule
  \end{tabular*}
\end{table}

\subsection{Soft Airframe in Nominal Free-Flight}
\label{sec:exp_cost}

In this experiment, we quantify the cost of the compliance in nominal contact-free flight, in terms of tracking accuracy and power consumption.
We compare the optimized vehicle against a rigid twin: the identical airframe with the two elastic side joints replaced by rigid connections of the same geometry and with the same weight.
Both configurations fly a $2.4$\,m-diameter circular trajectory at $1.5$\,m/s for $5$\,min per flight.
Each configuration flew three flights with the same three battery packs (six flights in total), with the same NMPC in both cases.

\begin{figure}[t]
  \centering
  \includegraphics[width=\columnwidth]{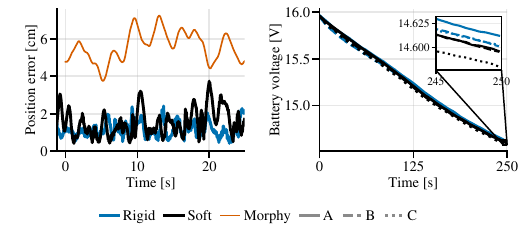}
  \caption{Nominal-flight cost of the passive joints on a $2.4$\,m-diameter circular trajectory at $1.5$\,m/s.
  Left: position error for soft-arm and rigid-arm flights and Morphy~\cite{morphy}.
  Right: battery voltage during the six $250$\,s comparison windows, where line color identifies the arm configuration and line style identifies the battery pack.}
  \label{fig:nominal_flight_cost}
\end{figure}

Figure~\ref{fig:nominal_flight_cost} shows tracking error and battery voltage over time.
The soft arms increase position RMSE from $1.5 \pm 0.4$ to $1.7 \pm 0.5$\,cm.
Both configurations track considerably better than Morphy~\cite{morphy}, whose position RMSE is $5.9 \pm 0.5$\,cm.
Flights took off at different states of charge, so to make discharge rates comparable, each voltage trace was aligned to start at 15.96 V and cut to a shared $250$\,s window.
With soft arms, the mean electrical power (about $106$\,W) is $0.4$--$0.9\%$ higher than with rigid arms, and the final voltage differs by at most $0.022$\,V.
Thus, the compliance that enables traversal of $\minimumGapWidth$\,cm gaps costs only $2$\,mm of additional position RMSE and less than one percent of additional electrical power in nominal flight.

\subsection{Sim2real Transfer}
\label{sec:objective_transfer}

\begin{figure}[t]
  \centering
  \includegraphics[width=\columnwidth]{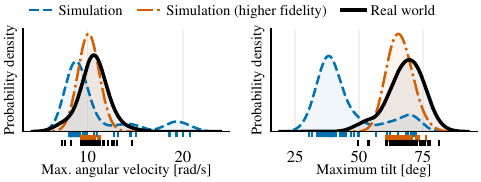}
  \caption{Left: maximum angular velocity norm.
  Right: maximum tilt.
  Both are measured over the first $0.2$\,s after crossing the gap plane at a $20$\,cm opening.}
  \label{fig:traversal_distributions}
\end{figure}

We further assess sim2real transfer by comparing the maximum angular velocity and tilt during the first $0.2$\,s after crossing a $20$\,cm gap in $50$ simulated and $50$ real-world traversals (Figure~\ref{fig:traversal_distributions}).
For the final optimized design, the simulation used for optimization underestimates the median maximum angular velocity by $15\%$ and tilt by $42\%$.
We attribute this mainly to differences in mass, inertia, and motor response, as the vehicle is $20\%$ heavier than the simulated one, its roll, pitch and yaw inertias differ by $13\%,18\%$ and $10\%$ respectively, and its motors respond to speed commands with a delay of about $14$\,ms that the simulation does not model.
Repeating the $50$ simulated traversals with the mass and inertia of the built vehicle and with motor responses fitted to the real rotor-speed logs reduces these gaps to $6\%$ and $4\%$ (higher-fidelity simulation in Figure~\ref{fig:traversal_distributions}).
Despite the discrepancy in the modeled dynamics during optimization, the proposed methodology guided the discovery of an effective design for gap traversal.  
More accurate simulation could potentially enable designs to achieve even better real-world performance.

\section{Conclusions}
\label{sec:conclusions}
In this work, we jointly optimized quadrotor geometry and passive joint response for gap-traversal success in closed-loop simulation.
The proposed method jointly selects geometry and softness directly based on the vehicle's ability to cross a gap and recover flight in simulation with an NMPC controller.
After optimization in simulation, we built a physical prototype that reduces its width from $\extendedWidth$ to $\foldedWidth$\,cm and traverses a $\minimumGapWidth \times 20$\,cm gap.
It can also pass through a $40$\,cm-long tube and recover after approach-angle errors up to $30^{\circ}$ at a $20$\,cm opening.
In nominal flight, passive joints increase position RMSE by $2$\,mm and electrical power by less than $1\%$ relative to a rigid twin.
More broadly, this work shows that closed-loop design optimization can produce airframe designs that fold passively to pass through tight gaps, while keeping nominal tracking accuracy and power consumption close to those of a rigid airframe.

\bibliographystyle{IEEEtran}
\bibliography{references}

\end{document}